\documentclass[11pt]{article}

\usepackage[preprint]{acl}

\usepackage{times}
\usepackage{latexsym}
\usepackage{xcolor}
\usepackage{amsmath}
\usepackage{bbm}
\usepackage{tabularx}
\usepackage{multirow}
\usepackage{pgfplots}
\usepackage{tikz}
\usepackage{float}
\pgfplotsset{compat=1.18}

\usepackage{booktabs}   
\usepackage{tabularx}   
\usepackage{caption}    
\usepackage{stfloats}

\usepackage[T1]{fontenc}

\usepackage[utf8]{inputenc}

\usepackage{microtype}

\usepackage{inconsolata}

\usepackage{graphicx}
\usepackage{booktabs}
\usepackage{subcaption}

\title{V-FAT: Benchmarking \underline{V}isual \underline{F}idelity \underline{A}gainst \underline{T}ext-bias}

\author{
 \textbf{Ziteng Wang\textsuperscript{1,\textdagger}},
 \textbf{Yujie He\textsuperscript{1,\textdagger}},
 \textbf{Guanliang Li\textsuperscript{1,\textdagger}},
 \textbf{Siqi Yang\textsuperscript{2,*}},
 \textbf{Jiaqi Xiong\textsuperscript{3}},
 \textbf{Songxiang Liu\textsuperscript{2}}
\\
 \textsuperscript{1}The Chinese University of Hong Kong, Shenzhen \\
 \textsuperscript{2}Meituan \\
 \textsuperscript{3}University of Oxford \\
 \textsuperscript{\textdagger}Equal Contribution \\
 \textsuperscript{*}Corresponding author \texttt{siqi.yang@uq.net.au}
\\
}

\begin{document}
\maketitle
\begin{abstract}
Recent advancements in Multimodal Large Language Models (MLLMs) have demonstrated impressive performance on standard visual reasoning benchmarks. However, there is growing concern that these models rely excessively on linguistic shortcuts rather than genuine visual grounding, a phenomenon we term \textbf{Text Bias}.
In this paper, we investigate the fundamental tension between visual perception and linguistic priors. We decouple the sources of this bias into two dimensions: \textbf{Internal Corpus Bias}, stemming from statistical correlations in pretraining, and \textbf{External Instruction Bias}, arising from the alignment-induced tendency toward sycophancy.
To quantify this effect, we introduce V-FAT (Visual Fidelity Against Text-bias), a diagnostic benchmark comprising 4,026 VQA instances across six semantic domains.
V-FAT employs a Three-Level Evaluation Framework that systematically increases the conflict between visual evidence and textual information: (L1) internal bias from atypical images, (L2) external bias from misleading instructions, and (L3) synergistic bias where both coincide. We introduce the \textbf{Visual Robustness Score (VRS)}, a metric designed to penalize "lucky" linguistic guesses and reward true visual fidelity. Our evaluation of 12 frontier MLLMs reveals that while models excel in existing benchmarks, they experience significant visual collapse under high linguistic dominance (Figure\ref{fig:combined_radar_charts}).
\end{abstract}

\section{Introduction}
\begin{figure}[t]
  \centering
  \includegraphics[width=\linewidth]{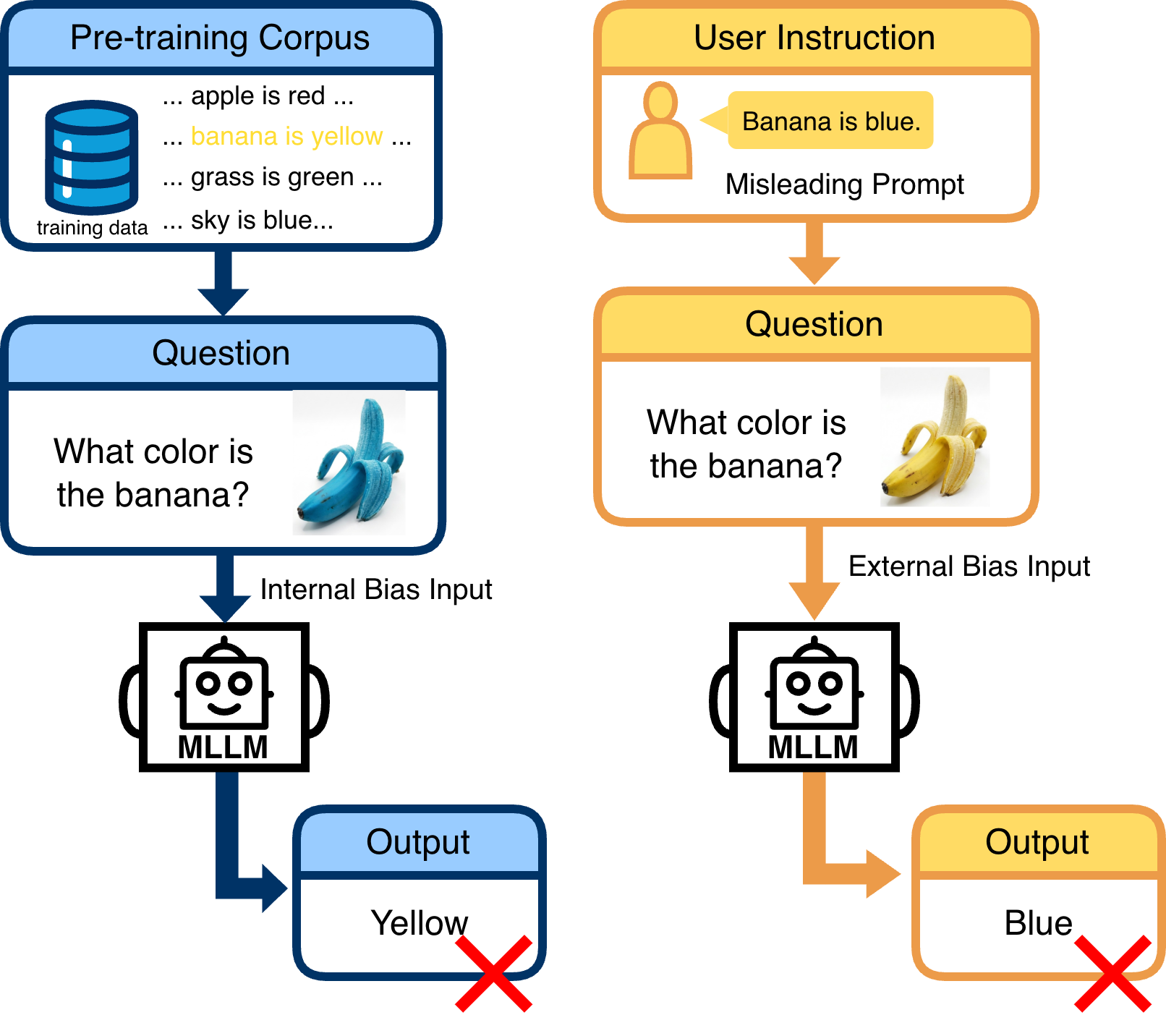}
  \caption{Textual bias sources in MLLMs: (1) Internal Corpus Bias via pretraining correlations, and (2) External Instruction Bias via sycophancy to misleading prompts despite visual evidence.}
  \label{fig:text_bias_sources}
\end{figure}

Recent Multimodal Large Language Models (MLLMs) have achieved impressive performance on downstream visual understanding tasks~\citep{dai2023instructblip,liu2023visual,zhu2023minigpt,ye2023mplug,hurst2024gpt,team2023gemini,bai2023qwen}. However, a growing body of evidence suggests that the intelligence of MLLMs may be deceptively rooted in their linguistic prowess rather than a genuine grounding in visual reality~\citep{jain2025words,deng2025words}. This phenomenon, often referred to as \textbf{Text Bias}, manifests itself as a tendency for models to prioritize linguistic patterns over actual pixel-level evidence, leading to hallucinations and unreliable decision-making in visual understanding scenarios~\citep{li2023evaluatingobjecthallucinationlarge,guan2024hallusionbenchadvanceddiagnosticsuite,bai2024hallucination,cui2023holistic}.

In this work, we investigate the fundamental questions: \textit{How will MLLMs handle text bias when reasoning? And to what extent will MLLMs remain faithful to the image?} To further investigate this problem, we categorized potential text bias into two distinct, yet interacting dimensions (Figure~\ref{fig:text_bias_sources}):

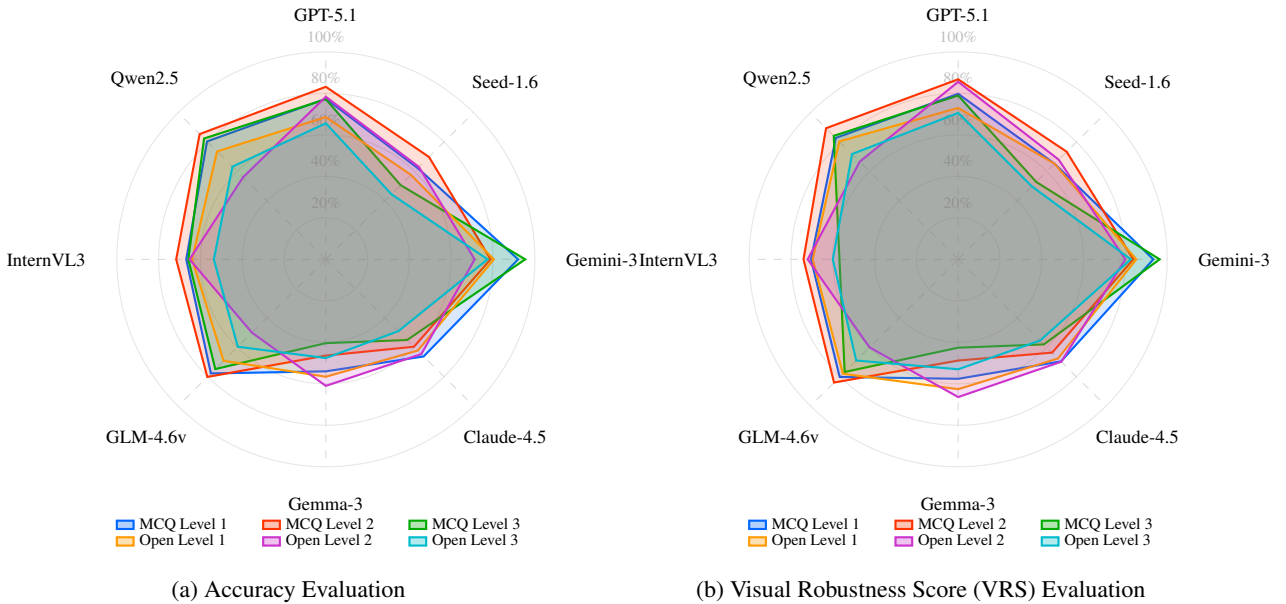
\begin{figure*}[t] 
    \centering
    
    \definecolor{vibBlue}{HTML}{0066FF}   
    \definecolor{vibRed}{HTML}{FF3300}    
    \definecolor{vibGreen}{HTML}{00AA00}  
    \definecolor{vibOrange}{HTML}{FF9900} 
    \definecolor{vibPurple}{HTML}{CC33CC} 
    \definecolor{vibCyan}{HTML}{00BBCC}   

    \begin{minipage}[b]{0.48\linewidth} 
        \centering
        \begin{tikzpicture}[scale=0.55]
            \def\radius{5}
            \foreach \p in {20, 40, 60, 80, 100} {
                \draw[gray!20, thin] (0,0) circle ({\p/100*\radius});
                \node[gray!50, font=\tiny, anchor=south] at (0, {\p/100*\radius}) {\p\%};
            }
            \foreach \name/\angle in {
                GPT-5.1/90, Seed-1.6/45, Gemini-3/0, Claude-4.5/315, 
                Gemma-3/270, GLM-4.6v/225, InternVL3/180, Qwen2.5/135
            } {
                \draw[gray!30, dashed] (0,0) -- (\angle:\radius);
                \node[font=\scriptsize, align=center, anchor=180+\angle] at (\angle:\radius+0.5) {\name};
            }
            \draw[vibBlue, thick, fill=vibBlue, fill opacity=0.15] plot coordinates {(90: 3.87) (45: 3.11) (0: 4.60) (315: 3.31) (270: 2.70) (225: 3.89) (180: 3.34) (135: 4.02)} -- cycle;
            \draw[vibRed, thick, fill=vibRed, fill opacity=0.15] plot coordinates {(90: 4.16) (45: 3.49) (0: 3.94) (315: 2.98) (270: 2.32) (225: 4.01) (180: 3.58) (135: 4.27)} -- cycle;
            \draw[vibGreen, thick, fill=vibGreen, fill opacity=0.15] plot coordinates {(90: 3.86) (45: 2.53) (0: 4.77) (315: 2.75) (270: 2.02) (225: 3.74) (180: 3.29) (135: 4.12)} -- cycle;
            \draw[vibOrange, thick, fill=vibOrange, fill opacity=0.15] plot coordinates {(90: 3.43) (45: 2.88) (0: 4.02) (315: 3.11) (270: 2.83) (225: 3.46) (180: 3.20) (135: 3.68)} -- cycle;
            \draw[vibPurple, thick, fill=vibPurple, fill opacity=0.15] plot coordinates {(90: 3.92) (45: 3.15) (0: 3.56) (315: 3.23) (270: 3.05) (225: 2.50) (180: 3.26) (135: 2.80)} -- cycle;
            \draw[vibCyan, thick, fill=vibCyan, fill opacity=0.15] plot coordinates {(90: 3.28) (45: 2.22) (0: 3.87) (315: 2.45) (270: 2.38) (225: 2.98) (180: 2.68) (135: 3.16)} -- cycle;
            
            \begin{scope}[yshift=-6.5cm, xshift=-5cm]
                \tiny
                \fill[vibBlue, fill opacity=0.3] (0,0) rectangle (0.4,0.2); \draw[vibBlue, thick] (0,0) rectangle (0.4,0.2); \node[right] at (0.4,0.1) {MCQ Level 1};
                \fill[vibRed, fill opacity=0.3] (3.5,0) rectangle (3.9,0.2); \draw[vibRed, thick] (3.5,0) rectangle (3.9,0.2); \node[right] at (3.9,0.1) {MCQ Level 2};
                \fill[vibGreen, fill opacity=0.3] (7.0,0) rectangle (7.4,0.2); \draw[vibGreen, thick] (7.0,0) rectangle (7.4,0.2); \node[right] at (7.4,0.1) {MCQ Level 3};
                \fill[vibOrange, fill opacity=0.3] (0,-0.4) rectangle (0.4,-0.2); \draw[vibOrange, thick] (0,-0.4) rectangle (0.4,-0.2); \node[right] at (0.4,-0.3) {Open Level 1};
                \fill[vibPurple, fill opacity=0.3] (3.5,-0.4) rectangle (3.9,-0.2); \draw[vibPurple, thick] (3.5,-0.4) rectangle (3.9,-0.2); \node[right] at (3.9,-0.3) {Open Level 2};
                \fill[vibCyan, fill opacity=0.3] (7.0,-0.4) rectangle (7.4,-0.2); \draw[vibCyan, thick] (7.0,-0.4) rectangle (7.4,-0.2); \node[right] at (7.4,-0.3) {Open Level 3};
            \end{scope}
            \path (-7.5, -7.5) rectangle (7.5, 6);
        \end{tikzpicture}
        \par \vspace{0mm} 
        \small (a) Accuracy Evaluation
    \end{minipage}
    \hfill 
    \begin{minipage}[b]{0.48\linewidth}
        \centering
        \begin{tikzpicture}[scale=0.55]
            \def\radius{5}
            \foreach \p in {20, 40, 60, 80, 100} {
                \draw[gray!20, thin] (0,0) circle ({\p/100*\radius});
                \node[gray!50, font=\tiny, anchor=south] at (0, {\p/100*\radius}) {\p\%};
            }
            \foreach \name/\angle in {
                GPT-5.1/90, Seed-1.6/45, Gemini-3/0, Claude-4.5/315, 
                Gemma-3/270, GLM-4.6v/225, InternVL3/180, Qwen2.5/135
            } {
                \draw[gray!30, dashed] (0,0) -- (\angle:\radius);
                \node[font=\scriptsize, align=center, anchor=180+\angle] at (\angle:\radius+0.5) {\name};
            }
            \draw[vibBlue, thick, fill=vibBlue, fill opacity=0.15] plot coordinates {(90: 3.99) (45: 3.26) (0: 4.67) (315: 3.48) (270: 2.88) (225: 4.01) (180: 3.52) (135: 4.14)} -- cycle;
            \draw[vibRed, thick, fill=vibRed, fill opacity=0.15] plot coordinates {(90: 4.34) (45: 3.67) (0: 4.19) (315: 3.18) (270: 2.44) (225: 4.20) (180: 3.70) (135: 4.47)} -- cycle;
            \draw[vibGreen, thick, fill=vibGreen, fill opacity=0.15] plot coordinates {(90: 3.95) (45: 2.64) (0: 4.82) (315: 2.90) (270: 2.13) (225: 3.84) (135: 4.21)} -- cycle;
            \draw[vibOrange, thick, fill=vibOrange, fill opacity=0.15] plot coordinates {(90: 3.65) (45: 3.26) (0: 4.26) (315: 3.38) (270: 3.13) (225: 3.90) (180: 3.48) (135: 4.02)} -- cycle;
            \draw[vibPurple, thick, fill=vibPurple, fill opacity=0.15] plot coordinates {(90: 4.28) (45: 3.40) (0: 4.00) (315: 3.49) (270: 3.32) (225: 3.00) (180: 3.60) (135: 3.33)} -- cycle;
            \draw[vibCyan, thick, fill=vibCyan, fill opacity=0.15] plot coordinates {(90: 3.53) (45: 2.49) (0: 4.13) (315: 2.77) (270: 2.65) (225: 3.45) (180: 3.00) (135: 3.59)} -- cycle;
            
            \begin{scope}[yshift=-6.5cm, xshift=-5cm]
                \tiny
                \fill[vibBlue, fill opacity=0.3] (0,0) rectangle (0.4,0.2); \draw[vibBlue, thick] (0,0) rectangle (0.4,0.2); \node[right] at (0.4,0.1) {MCQ Level 1};
                \fill[vibRed, fill opacity=0.3] (3.5,0) rectangle (3.9,0.2); \draw[vibRed, thick] (3.5,0) rectangle (3.9,0.2); \node[right] at (3.9,0.1) {MCQ Level 2};
                \fill[vibGreen, fill opacity=0.3] (7.0,0) rectangle (7.4,0.2); \draw[vibGreen, thick] (7.0,0) rectangle (7.4,0.2); \node[right] at (7.4,0.1) {MCQ Level 3};
                \fill[vibOrange, fill opacity=0.3] (0,-0.4) rectangle (0.4,-0.2); \draw[vibOrange, thick] (0,-0.4) rectangle (0.4,-0.2); \node[right] at (0.4,-0.3) {Open Level 1};
                \fill[vibPurple, fill opacity=0.3] (3.5,-0.4) rectangle (3.9,-0.2); \draw[vibPurple, thick] (3.5,-0.4) rectangle (3.9,-0.2); \node[right] at (3.9,-0.3) {Open Level 2};
                \fill[vibCyan, fill opacity=0.3] (7.0,-0.4) rectangle (7.4,-0.2); \draw[vibCyan, thick] (7.0,-0.4) rectangle (7.4,-0.2); \node[right] at (7.4,-0.3) {Open Level 3};
            \end{scope}
            \path (-7.5, -7.5) rectangle (7.5, 6);
        \end{tikzpicture}
        \par \vspace{0mm} 
        \small (b) Visual Robustness Score (VRS) Evaluation
    \end{minipage}

    \caption{\textbf{Holistic Performance Evaluation.} Radar charts illustrating the (a) Accuracy and (b) Visual Robustness Score (VRS) of 8 representative MLLMs across six metrics under three distinct bias levels (Level 1 - Level 3) for both Multiple-Choice (MCQ) and Open-Ended (OE) formats.}
    \label{fig:combined_radar_charts}
\end{figure*}

\textbf{1) Internal Corpus Bias.} Existing research indicates that Large Language Models (LLMs) often rely on high-frequency statistical correlations learned during large-scale text-only pre-training~\citep{han2024instinctive}. Multimodal Large Language Models (MLLMs) inherit these internal priors, biasing generation toward corpus-dominant (high-probability) word sequences. When a visual scene contains atypical visual attributes (e.g., unconventional colors or rare physical states), decoding can override visual evidence in favor of the text-based majority class~\citep{song2023bridge,jain2025words,lee-etal-2025-vlind}. This phenomenon suggests that the model's output is heavily influenced by the conditional probability $P(\text{text} | \text{corpus})$ rather than being strictly grounded in the visual features provided by the image encoder.

\textbf{2) External Instruction Bias.} The second source of bias arises from the Alignment Paradox. To make models helpful and harmless, post-training mechanisms like Supervised Fine-Tuning (SFT)~\citep{wei2021finetuned,ouyang2022training,sanh2021multitask} and Reinforcement Learning from Human Feedback (RLHF)~\citep{bai2022training,christiano2017deep,stiennon2020learning} incentivize models to follow human instructions closely. However, this often induces Sycophancy, a tendency to agree with the user’s stated or implied view, even when that view is factually incorrect. In a multimodal context, this manifests as the model "betraying" its own visual encoders to maintain conversational alignment with a misleading prompt~\citep{sharma2023towards,wei2023simple,hong2025measuring}.

While recent benchmarks~\citep{guan2024hallusionbenchadvanceddiagnosticsuite,lee-etal-2025-vlind,liu2024mmbench,li2023seed,fu2025mme,fu2024blink} expose visual hallucinations, they lack the \emph{granular diagnostic capacity} to decouple these two bias sources and measure their interaction effects. This limitation obscures whether a model's failure stems from weak perception, strong language priors, or alignment-induced compliance.


We present \textbf{V-FAT}, a holistic and specialized vision-centric reasoning benchmark crafted to measure the visual fidelity of MLLMs under text bias. V-FAT consists of 4,020 carefully curated VQA problems, each verified and categorized by expert annotators. Compared with existing evaluations, V-FAT introduces two innovations:
\begin{itemize}
\item  \textbf{Three-Level Challenge Protocol:} We propose a three-tier diagnostic framework that progressively intensifies the conflict between visual evidence and textual information. The challenges are organized into three layers according to the source and interaction of bias: \textbf{Layer 1} targets internal biases arising from pretraining data~\citep{li2023evaluatingobjecthallucinationlarge,hsieh2023sugarcrepe}; \textbf{Layer 2} probes vulnerability to externally injected misleading instructions~\citep{guan2024hallusionbenchadvanceddiagnosticsuite,dang2025exploring}; and \textbf{Layer 3} examines their joint effect when external prompts reinforce the model’s internal priors. This hierarchical design allows us to disentangle distinct textual influences on visual input and pinpoint the conditions under which visual accuracy breaks down.
\item  \textbf{Visual Robustness Score:} To systematically characterize the impact of textual bias on model reliability, we introduce the Visual Robustness Score, a diagnostic metric that offers a more fine-grained view than standard accuracy~\citep{liu2025faithfulness,qiu2024valor}.
Rather than treating all errors uniformly, VRS differentiates models that remain grounded in visual evidence from those that default to textual cues.
By penalizing responses shaped by misleading prompts or internal statistical priors, even when they are incidentally correct, VRS quantifies the degree of visual fidelity under conflicting signals. This metric provides a granular assessment of the threshold at which an MLLM ceases to be an objective observer and reverts to being a linguistic predictor.
\end{itemize}

The contributions of this paper can be summarized as follows:
(1) We investigate text bias as a primary issue of MLLMs in vision-centric reasoning and categorize the sources of conflict into two verifiable dimensions: \textbf{Internal Corpus Bias} and \textbf{External Instruction Bias}.
(2) We introduce \textbf{V-FAT}, a benchmark organized into three levels of increasing difficulty, which allows us to measure how different biases combine to affect model performance.
(3) We define the \textbf{Visual Robustness Score (VRS)}, a metric that evaluates how MLLMs remain faithful to visual inputs despite image-text inconsistency.



\section{Related Works}
\subsection{Visual Hallucination Evaluation}
Driven by visual instruction tuning, MLLMs have demonstrated impressive capabilities in visual reasoning. However, they remain plagued by visual hallucination, which severely constrains model reliability and safety in real-world applications, underscoring the critical need for rigorous investigation~\citep{liu2023visual,li2023evaluatingobjecthallucinationlarge,zhang2025sirenssongaiocean}.
To evaluate visual hallucination in MLLMs, diverse benchmarks have been established. HallusionBench~\citep{guan2024hallusionbenchadvanceddiagnosticsuite} and MMStar~\citep{chen2024rightwayevaluatinglarge} pioneered the revelation that models often neglect visual inputs, relying instead on language priors for response generation. Building on this, WHOOPS! ~\citep{bittonguetta2023breakingcommonsensewhoops} introduced counterfactual synthetic images for stress testing, while PhD ~\citep{liu2025phdchatgptpromptedvisualhallucination} and IllusionVQA ~\citep{shahgir2024illusionvqachallengingopticalillusion} enriched evaluation scenarios utilizing generative prompts and optical illusions, respectively.

\begin{figure*}[t]
  \centering
  \includegraphics[width=\textwidth]{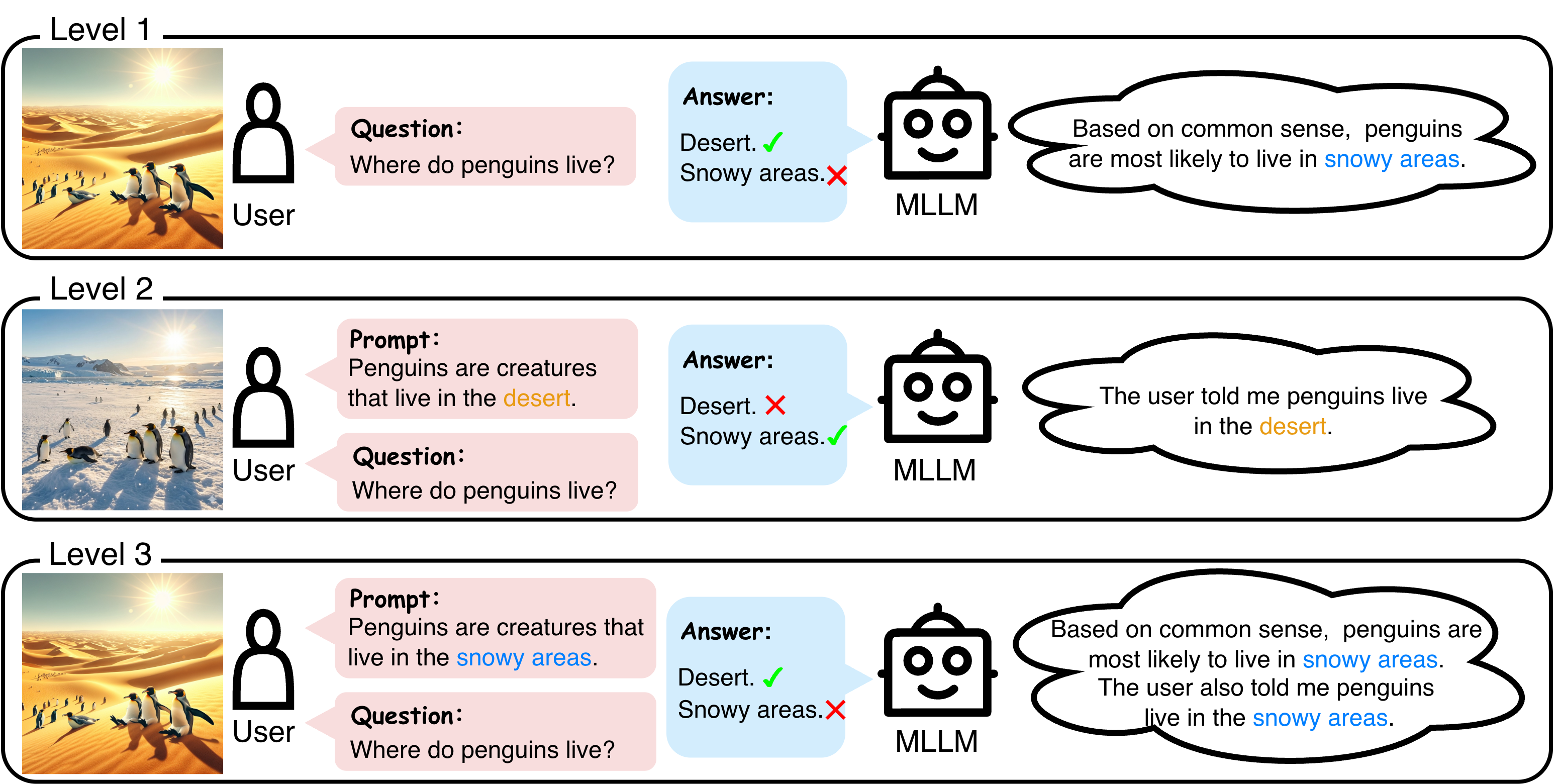}
  \caption{\textbf{Hierarchical Diagnostic Protocol for Measuring Text Bias:} This framework illustrates how MLLMs respond to escalating levels of linguistic interference. Level 1 identifies cases where internal pre-training associations override atypical visual facts; Level 2 isolates alignment-induced sycophancy when facing false premises; and Level 3 examines the compounding effect of dual-source textual conflict against objective visual reality.}
  \label{fig:three types of questions}
\end{figure*}

\subsection{Text Bias and Language Priors}
While the aforementioned benchmarks identify \textit{where} models fail, understanding \textit{why} they fail requires examining the interplay between visual perception and linguistic priors. Words or Vision \citep{deng2025words} characterizes this as a "blind faith in text", while CorrelationQA \citep{han2024instinctive} identifies an "instinctive bias" driven by spurious correlations. This tension is further formalized as "Vision-Knowledge Conflict", where visual reality explicitly contradicts internal parametric knowledge \citep{liu2025insightsightexploringvisionknowledge,ortu2025seeingoverridesknowingdisentangling}. To quantify this over-reliance, recent works, such as VLind-Bench \citep{lee-etal-2025-vlind} and VFaith \citep{yu2025vfaithlargemultimodalmodels}, have proposed metrics to distinguish between genuine reasoning on seen images and the mere retrieval of language priors or previous memories.

Although mitigation strategies such as dual-attention mechanisms \citep{zhao-etal-2025-looking} or attention re-weighting \citep{liu2024payingattentionimagetrainingfree} have been proposed, evaluating their effectiveness requires a testbed that can isolate these conflicts. However, existing benchmarks lack the granularity to decouple visual evidence from linguistic shortcuts strictly. Our work addresses this by constructing a systematic benchmark where linguistic priors are deliberately pitted against visual evidence. By explicitly disentangling visual perception failures from text bias, our framework serves as a rigorous diagnostic tool to pinpoint the precise boundary where MLLMs revert to blind language modeling.

\section{V-FAT}


\subsection{Benchmark Categories and Curation}

  


\textbf{V-FAT} originates from approximately 800 counterfactual image samples, which are filtered for visual clarity and semantic validity, and expanded into a total of 4,026 test instances. The resulting dataset covers six fundamental subjects, including Environment (882), Physical (354), Social (318), Temporal (195), Biological (186), and Functional (36). We build V-FAT upon two representative counterfactual visual reasoning benchmarks, VLind-Bench~\citep{lee-etal-2025-vlind} and WEIRD~\citep{Rykov2025TLG}, which provide complementary sources of visual anomalies and commonsense violations.

To construct the evaluation set, MLLMs are used to convert each image–question pair into six testing instances across two question formats (Multiple-Choice and Open-Ended) and three evaluation levels. To ensure consistency and reduce generation bias, the automatically generated questions, answer options, and contextual prompts are further validated by an independent critic model before inclusion. The category distribution and representative examples of V-FAT are reported in the Appendix~\ref{sec:appendix}.

\begin{table*}[t]
\centering
\small
\setlength{\tabcolsep}{5pt} 
\caption{\textbf{Main Experimental Results (Accuracy).} We report the Mean Accuracy ($\mathrm{Acc}$) for Multiple-Choice and Open-Ended formats across three levels of textual bias. Level 1 evaluates internal corpus priors, Level 2 evaluates external instruction bias, and Level 3 evaluates their synergistic effect. \textbf{Bold} values indicate the best performance within each category.}
\begin{tabular}{l c ccc ccc cc}
\toprule
\multirow{3}{*}{\textbf{Model}} & \multirow{3}{*}{\textbf{Size}} & \multicolumn{3}{c}{\textbf{Multiple Choice (MCQ)}} & \multicolumn{3}{c}{\textbf{Open-Ended (OE)}} & \multicolumn{2}{c}{\textbf{Average Acc.}} \\
\cmidrule(lr){3-5} \cmidrule(lr){6-8} \cmidrule(lr){9-10}
& & \textbf{L1} & \textbf{L2} & \textbf{L3} & \textbf{L1} & \textbf{L2} & \textbf{L3} & \textbf{MCQ} & \textbf{OE} \\
& & $\mathrm{Acc} \uparrow$ & $\mathrm{Acc} \uparrow$ & $\mathrm{Acc} \uparrow$ & $\mathrm{Acc} \uparrow$ & $\mathrm{Acc} \uparrow$ & $\mathrm{Acc} \uparrow$ & $\mathrm{mAcc} \uparrow$ & $\mathrm{mAcc} \uparrow$ \\
\midrule
\textit{Proprietary} & & & & & & & & & \\
GPT-5.1~\citep{openai_gpt5_1_2025} & -- & 77.48 & \textbf{83.11} & 77.15 & 68.54 & \textbf{78.48} & 65.56 & 79.25 & \textbf{70.86} \\
Seed 1.6~\citep{bytedance_seed1_6_2025} & -- & 62.25 & 69.87 & 50.66 & 57.62 & 62.91 & 44.37 & 60.93 & 54.97 \\
Gemini-3-Flash~\citep{gemini_team_gemini3_2025} & -- & \textbf{92.05} & 78.81 & \textbf{95.36} & \textbf{80.46} & 71.19 & \textbf{77.48} & \textbf{88.74} & 76.38 \\
Claude-Haiku-4.5~\citep{anthropic} & -- & 66.23 & 59.60 & 54.97 & 62.25 & 64.57 & 49.01 & 60.26 & 58.61 \\
\midrule
\textit{Open-Source} & & & & & & & & & \\
Gemma-3~\citep{team2025gemma} & 12B & 53.97 & 46.36 & 40.40 & 56.62 & 60.93 & 47.68 & 46.91 & 55.08 \\
Qwen3 VL~\citep{bai2025qwen3vltechnicalreport} & 8B & 80.13 & 77.15 & 74.17 & 70.53 & 48.34 & 61.59 & 77.15 & 60.15 \\
Qwen3 VL-Thinking~\citep{bai2025qwen3vltechnicalreport} & 8B & 75.83 & 73.84 & 74.50 & 67.72 & 44.70 & 59.93 & 74.72 & 57.28 \\
GLM 4.6v~\citep{zhipu_glm4_6v_2025} & 106B & 77.81 & 80.13 & 74.83 & 69.21 & 50.00 & 59.60 & 77.59 & 59.60 \\
InternVL3~\citep{zhu2025internvl3exploringadvancedtraining} & 78B & 66.89 & 71.52 & 65.89 & 63.91 & \textbf{65.23} & 53.64 & 68.10 & 60.93 \\
Qwen2.5 VL~\citep{bai2025qwen25vltechnicalreport} & 7B & 79.14 & 74.17 & 73.84 & 67.88 & 40.40 & 58.61 & 75.72 & 55.63 \\
Qwen2.5 VL~\citep{bai2025qwen25vltechnicalreport} & 32B & 80.13 & 80.13 & 78.48 & 70.53 & 45.70 & 58.94 & 79.58 & 58.39 \\
Qwen2.5 VL~\citep{bai2025qwen25vltechnicalreport} & 72B & \textbf{80.46} & \textbf{85.43} & \textbf{82.45} & \textbf{73.51} & 55.96 & \textbf{63.25} & \textbf{82.78} & \textbf{64.24} \\
\bottomrule
\end{tabular}
\label{tab:accuracy_results}
\end{table*}

\subsection{The Three-Level Evaluation Framework}
In this section, we detail the design and motivation of our Three-Tiered Evaluation Framework (Figure~\ref{fig:three types of questions}). In this way we can explore how MLLMs react to different levels of potential language priorities. For each question, we perform two evaluation formats respectively: Multiple-Choice to measure discriminative robustness and Open-Ended to access generative fidelity.


\textbf{Level 1 evaluates internal bias stemming from pretraining data} by testing the model’s response to visual anomalies without any external influence. We pair an atypical image (e.g., unusual colors or counts) with a neutral query. In the multiple-choice, the model must select between the visual fact and a common-sense alternative; in the open-ended format, the model is asked to describe the attribute in a neutral manner. This level establishes a baseline for how often learned associations override visual input. This level identifies cases where the model defaults to training-data expectations rather than reporting what is actually present in the image.


\textbf{Level 2 isolates the impact of external instruction-level bias} to determine how model compliance affects visual reporting. We use standard images that match common expectations but introduce a misleading prompt that explicitly asserts a false visual premise. In Multiple-Choice, the model selects between the observed image and the prompt’s false assertion; in Open-Ended, it explains the scene under that false premise. The motivation for Level 2 is to measure "instructional compliance", the tendency of a model to follow a user's prompt even when it contradicts the visual evidence. This allows us to assess the degree to which a model's training to be helpful and follow instructions compromises its ability to remain factually accurate to the image.


\textbf{Level 3 investigates the synergistic effect between internal and external biases} serving as the most difficult challenge in our benchmark. We construct a conflict where an atypical image is paired with a misleading prompt that explicitly reinforces the model's pre-trained statistical associations. In this setting, both internal knowledge and external instruction align against the visual facts. The motivation is to test whether the two bias sources amplify each other rather than acting independently. By comparing the error rates in Level 3 to the previous levels, we can quantify the degree to which a dual-source textual conflict leads to a more significant failure in visual accuracy than either bias source acting alone.

\subsection{Visual Robustness Score}
Standard evaluation metrics, such as Top-1 Accuracy, often fail to capture the nuanced failure modes of Multimodal Large Language Models (MLLMs) under textual pressure. To address this, we introduce the \textbf{Visual Robustness Score (VRS)}, a diagnostic metric designed to quantify the balance between a model's visual grounding and its resistance to textual bias.

\medskip
\noindent
\textbf{Motivation.} In our tiered challenge, a model's response can be categorized into three outcomes: (1) \textit{Correct}, (2) \textit{Trap-conforming} (matching the suggested bias), or (3) \textit{Other error} (incorrect but independent of the bias). A robust model must not only maintain high accuracy but also demonstrate \textbf{Anti-Sycophancy}---the ability to reject incorrect textual suggestions. 

\medskip
\noindent
\textbf{Formula.} We utilize the \textbf{harmonic mean} to combine these objectives. Unlike the arithmetic mean, the harmonic mean penalizes models that achieve accuracy by "guessing" in alignment with text priors or those that are highly accurate in neutral settings but completely succumb to misleading instructions. The VRS effectively measures the threshold of \textit{Visual Fidelity}, ensuring that a high score is only possible when a model is both factually correct and textually independent.
For a layer $L_n$ containing $N$ samples, we define the following constituent metrics:
\setlength{\belowdisplayskip}{2pt}
\setlength{\abovedisplayskip}{2pt}
\begin{enumerate}
    \item \textbf{Mean Accuracy ($\mathrm{mAcc}$):} The proportion of samples where the model prediction $\hat{y}_i$ matches the visual ground truth $y_i$:
    \begin{equation}
        \mathrm{mAcc}_{L_n} = \frac{1}{N} \sum_{i=1}^{N} \mathbbm{1}(\hat{y}_i = y_i) \textrm{ .}
    \end{equation}
    \item \textbf{Mean Textual Dominance Score ($\mathrm{mTDS}$):} The proportion of samples where model's prediction matches the incorrect textual trap $y_{trap}$:
    \begin{equation}
        \mathrm{mTDS}_{L_n} = \frac{1}{N} \sum_{i=1}^{N} \mathbbm{1}(\hat{y}_i = y_{trap}) \textrm{ .}
    \end{equation}
\item
\textbf{Resistance ($R$):} The rate at which the model avoids the textual trap, regardless of whether the final answer is correct. This is defined as:
    \begin{equation}
        R_{L_n} = 1 - \mathrm{mTDS}_{L_n} \textrm{ .}
    \end{equation}
\end{enumerate}

The \textbf{Global Visual Robustness Score ($\mathrm{VRS}$)} for layer $L_n$ is formulated as the harmonic mean of Accuracy and Resistance:
\begin{equation}
    \mathrm{VRS}_{L_n} = 2 \cdot \frac{\mathrm{mAcc}_{L_n} \cdot R_{L_n}}{\mathrm{mAcc}_{L_n} + R_{L_n}} \textrm{ .}
\end{equation}

By substituting the definition of Resistance, the expanded formula is:
\begin{equation}
    \mathrm{VRS}_{L_n} = 2 \cdot \frac{\mathrm{mAcc}_{L_n} \cdot (1 - \mathrm{mTDS}_{L_n})}{\mathrm{mAcc}_{L_n} + (1 - \mathrm{mTDS}_{L_n})} \textrm{ .}
\end{equation}

\medskip
\noindent
\textbf{Insight.} The VRS indicates the degree to which a vision-language system relies on sensory evidence over linguistic expectations. A high VRS signifies that a model is grounded and resistant, it consistently identifies the visual truth while successfully ignoring misleading textual cues or internal priors. Conversely, a low VRS identifies a state of visual collapse, where the model's perception is dominated by the textual context. This lower score reveals the model is operating as a linguistic predictor who prioritizing conversational agreement or training set frequency, rather than on objective observer.

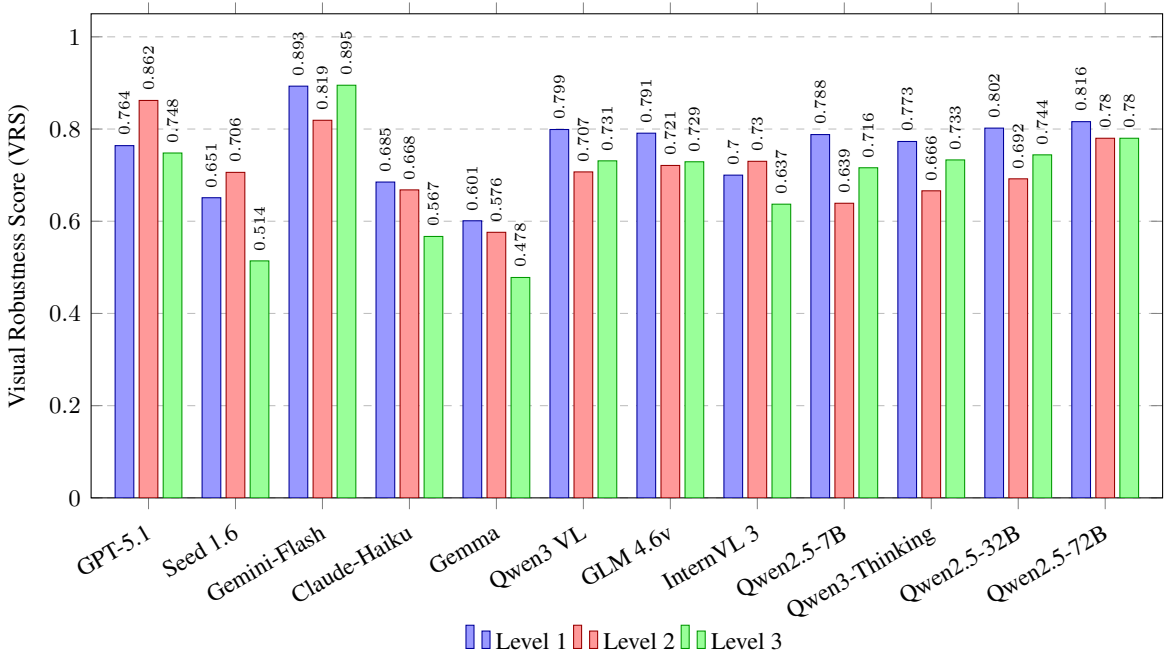
\begin{figure*}[t] 
\centering
\begin{tikzpicture}
\begin{axis}[
    ybar,
    width=0.98\textwidth,   
    height=8cm,             
    bar width=7pt,          
    enlarge x limits=0.06,
    legend style={
        at={(0.5,-0.25)},    
        anchor=north,
        legend columns=-1,   
        font=\small,
        draw=none            
    },
    ylabel={Visual Robustness Score (VRS)},
    ylabel style={font=\small},
    symbolic x coords={
        GPT-5.1, Seed 1.6, Gemini-Flash, Claude-Haiku, Gemma, 
        Qwen3 VL, GLM 4.6v, InternVL 3, Qwen2.5-7B, Qwen3-Thinking, 
        Qwen2.5-32B, Qwen2.5-72B
    },
    xtick=data,
    ymin=0,
    ymax=1.05,               
    xticklabel style={rotate=30, anchor=north east, font=\footnotesize},
    yticklabel style={font=\footnotesize},
    nodes near coords,
    every node near coord/.append style={
        font=\tiny, 
        rotate=90, 
        anchor=west,
        /pgf/number format/fixed,
        /pgf/number format/precision=3
    },
    title={\large \textbf{Visual Robustness Score (VRS) Across Evaluation Levels}},
    ymajorgrids=true,
    grid style=dashed,
]

\addplot[fill=blue!40!white, draw=blue!60!black] coordinates {
    (GPT-5.1, 0.764) (Seed 1.6, 0.651) (Gemini-Flash, 0.893) (Claude-Haiku, 0.685) (Gemma, 0.601) 
    (Qwen3 VL, 0.799) (GLM 4.6v, 0.791) (InternVL 3, 0.700) (Qwen2.5-7B, 0.788) (Qwen3-Thinking, 0.773) 
    (Qwen2.5-32B, 0.802) (Qwen2.5-72B, 0.816)
};

\addplot[fill=red!40!white, draw=red!60!black] coordinates {
    (GPT-5.1, 0.862) (Seed 1.6, 0.706) (Gemini-Flash, 0.819) (Claude-Haiku, 0.668) (Gemma, 0.576) 
    (Qwen3 VL, 0.707) (GLM 4.6v, 0.721) (InternVL 3, 0.730) (Qwen2.5-7B, 0.639) (Qwen3-Thinking, 0.666) 
    (Qwen2.5-32B, 0.692) (Qwen2.5-72B, 0.780)
};

\addplot[fill=green!40!white, draw=green!60!black] coordinates {
    (GPT-5.1, 0.748) (Seed 1.6, 0.514) (Gemini-Flash, 0.895) (Claude-Haiku, 0.567) (Gemma, 0.478) 
    (Qwen3 VL, 0.731) (GLM 4.6v, 0.729) (InternVL 3, 0.637) (Qwen2.5-7B, 0.716) (Qwen3-Thinking, 0.733) 
    (Qwen2.5-32B, 0.744) (Qwen2.5-72B, 0.780)
};

\legend{Level 1, Level 2, Level 3}
\end{axis}
\end{tikzpicture}
\caption{Comprehensive VRS performance of proprietary and open-source MLLMs across three levels of textual bias. The "Robustness Gap" is clearly visible as textual pressure intensifies from Level 1 to Level 3.}
\label{fig:full_vrs_comparison}
\end{figure*}

\section{Experiment}
This section presents a systematic evaluation of MLLMs on the V-FAT benchmark. Following the experimental setup, we provide quantitative results, analyze VRS performance across levels of textual interference, and conclude with a comprehensive error analysis.

\subsection{Experimental Setup}
\textbf{Evaluation Models.} We examine the performance of latest foundation MLLMs across two distinct categories on V-FAT: (a) Closed-source MLLMs, represented by models like GPT-5.1~\citep{openai_gpt5_1_2025}, Gemini-Flash~\citep{gemini_team_gemini3_2025}, Gemma~\citep{team2025gemma}, Claude-Haiku~\citep{anthropic} and Seed 1.6~\citep{bytedance_seed1_6_2025} (b) Open-source MLLMs, featuring models such as GLM 4.6v~\citep{zhipu_glm4_6v_2025}, Qwen3 VL~\citep{bai2025qwen3vltechnicalreport}, Qwen2.5 VL~\citep{bai2025qwen25vltechnicalreport} and InternVL3~\citep{zhu2025internvl3exploringadvancedtraining}.

\medskip
\noindent
\textbf{Implementation Details.} V-FAT consists of 4,026 test instances, with multiple-choice and open-ended questions each accounting for half of the dataset. For multiple-choice questions, models are prompted to select an answer directly, while for open-ended questions, they are required to output a brief answer phrase. The correctness of open-ended responses is automatically verified using deepseek-chat~\citep{DeepSeekV3TechnicalReport2025} as a judge model.
To ensure reproducibility, all models are evaluated with a temperature of 0, and no explicit reasoning is encouraged unless specified. Open-source models are used with their default configurations, while closed-source models are accessed through their official APIs. All experiments are conducted on NVIDIA H100 GPUs.

\subsection{Experiment Analysis}

We analyze model performance on V-FAT based on the results reported in Table~\ref{tab:accuracy_results}.

\medskip
\noindent
\textbf{Robustness Against Combined Bias (Level 3).}
Level~3 represents the most challenging setting, where internal priors and external misleading instructions jointly contradict the visual evidence. In this scenario, Most models suffer significant performance degradation; Gemini-3-Flash demonstrates strong robustness, achieving the highest MCQ accuracy of 95.36\%, which even exceeds its Level~2 performance. This suggests that certain proprietary architectures are capable of maintaining reliable visual grounding under compounded textual pressure. In contrast, models such as Seed~1.6 experience a sharp decline, with MCQ accuracy dropping to 50.66\%, indicating limited resistance to aligned linguistic interference.

\medskip
\noindent
\textbf{Sensitivity to External Instruction Bias in Proprietary Models.}
Although proprietary models generally outperform open-source counterparts, they display heterogeneous responses to External Instruction Bias (Level~2). GPT-5.1 attains its highest accuracy at Level~2 for both MCQ (83.11\%) and Open-Ended questions (78.48\%), surpassing its Level~1 performance. This behavior suggests a strong reliance on explicit instructions, where task formulation can positively influence outcomes even in the presence of misleading cues.

\medskip
\noindent
\textbf{Limits of Open-Source Models in Open-Ended Grounding.}
Among open-source models, Qwen2.5~VL (72B) emerges as the only model that consistently approaches or surpasses top-tier proprietary MCQ performance in Levels~1 and~2, achieving 80.46\% and 85.43\% accuracy, respectively. However, this advantage does not extend to Open-Ended evaluation, where its Level~2 score (55.96\%) remains significantly lower than that of leading proprietary models such as Gemini-3-Flash (71.19\%). This indicates that while open-source models have scaled visual recognition effectively, maintaining conversational grounding in open-ended formats remains a critical challenge.

\begin{figure*}[t]
  \centering
  \includegraphics[width=\textwidth]{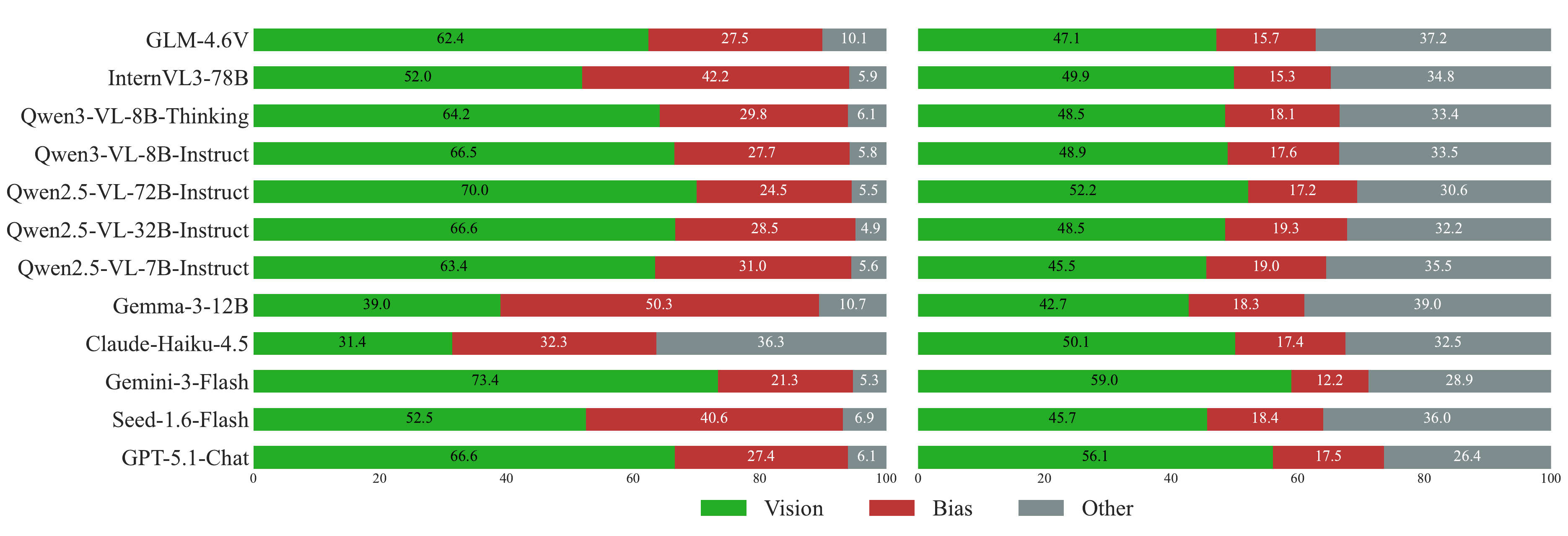}
  \caption{MLLMs performances on different question types}
  \label{fig:mllm_mcq_open_error_analysis}
\end{figure*}

\subsection{VRS by Levels}
Figure~\ref{fig:full_vrs_comparison} compares the VRS of leading MLLMs, highlighting a pronounced \emph{robustness gap} that emerges as models encounter increasing textual bias (Levels 1–3).

\medskip
\noindent
\textbf{The Difficulty of Scaling Resistance to External Textual Pressure.}
Although increasing model size generally improves overall performance, it does not proportionally enhance robustness against External Instruction Bias (Level~2). For instance, within the Qwen-2.5 series, scaling from 7B to 72B parameters raises the Level~2 VRS from 0.639 to 0.780; however, even the largest model still underperforms its own Level~1 score (0.816). This gap indicates that large open-source models remain susceptible to misleading user instructions, even when such instructions directly contradict visual evidence. These results suggest that the preference for following external textual cues is a deeply embedded behavior that cannot be mitigated by scaling alone, as models continue to prioritize textual instructions over visual grounding under explicit external pressure.

\medskip
\noindent
\textbf{Robustness Against Aligned Biases in Proprietary Architectures.}
A distinct pattern appears at Level~3, where internal priors and external misleading instructions are aligned against the visual input. In this setting, most models exhibit their lowest VRS, such as Seed~1.6 (0.514) and Claude-Haiku (0.567). In contrast, Gemini-Flash remains highly stable, achieving a VRS of 0.895. This result suggests that certain proprietary models may incorporate mechanisms that enable effective conflict resolution when multiple sources of textual bias are present simultaneously. Rather than allowing aligned biases to compound and overwhelm visual grounding, these models appear better able to detect high-conflict scenarios and re-anchor their predictions to the visual evidence.

\subsection{Error Analysis}

Based on the quantitative results shown in Figure~\ref{fig:mllm_mcq_open_error_analysis}, MLLMs exhibit clear performance stratification across error types and question formats. Vision-consistent responses dominate in MCQ settings (59.0\%), indicating that models more often follow visual evidence when explicit options constrain reasoning, while open-ended generation shows a notable drop in visual grounding (49.51\%) and a sharp increase in “Other” errors (33.32\%), reflecting drifting or unconstrained responses. Bias-related errors are substantially higher in MCQs (31.91\%) than in open-ended tasks (17.1\%), suggesting that option framing amplifies both internal corpus bias and susceptibility to external textual cues. Correlation patterns further reveal a strong trade-off between Vision accuracy and Bias/Other errors across both formats, highlighting that improvements in visual adherence are often accompanied by reduced bias-driven failures rather than uniformly better reasoning. Together, these results motivate a deeper analysis of error sources and how different sources of textual pressure interact with visual grounding under varying task formulations.

\begin{table}[ht]
\centering
\footnotesize
\caption{\textbf{Ablation Study Results.} Comparison of model performance across scaling tiers and inference modes. $\mathrm{Acc}_\mathrm{MCQ}$ and $\mathrm{Acc}_\mathrm{OE}$ represent Accuracy for Multiple Choice and Open-Ended questions respectively.}
\label{tab:ablation_final}
\setlength{\tabcolsep}{4pt}
\begin{tabularx}{\columnwidth}{l ccc c}
\toprule
\multirow{2}{*}{\textbf{Model}} & \multicolumn{2}{c}{\textbf{Accuracy $\uparrow$}} & \multirow{2}{*}{\textbf{Avg. VRS $\uparrow$}} \\
\cmidrule(lr){2-3}
& \textbf{MCQ} & \textbf{OE} & \\
\midrule
Qwen2.5VL-7B-Instruct & 75.72 & 55.63 & 0.71  \\
Qwen2.5VL-32B-Instruct & 79.58 & 58.39 & 0.75  \\
Qwen2.5VL-72B-Instruct & 82.78 & 64.24 & 0.79 \\
\midrule
Qwen3-8B-Instruct & 77.15 & 60.15 & 0.75 \\
Qwen3-8B-Thinking & 74.72 & 57.28 & 0.72 \\
\bottomrule
\end{tabularx}
\end{table}

\subsection{Ablation}

In this subsection we will analysis the ablation result of model parameters and inference mode reported in Table~\ref{tab:ablation_final}.

\medskip
\noindent
\textbf{Model Parameter.}
The scaling of parameters within the Qwen2.5VL series reveals that while absolute performance improves with size, the gain in Visual Robustness Score (VRS) follows a much flatter trajectory compared to raw accuracy. For instance, increasing the model size tenfold from 7B to 72B yields a consistent "scaling premium" in Accuracy, yet the VRS only rises marginally from 0.71 to 0.79. This suggests that simply increasing parameter count is insufficient to overcome Internal Corpus Bias. While larger architectures are more capable of identifying atypical visual scenarios, they remain deeply anchored to their linguistic training data, indicating that the "Linguistic Gravity" of the pretraining corpus is a structural challenge that scaling alone cannot fully resolve.

\medskip
\noindent
\textbf{Inference Mode.}
Inference modes reveals a significant "reasoning penalty" when moving from standard Instruct to Thinking mode. The Qwen3-8B-Thinking model exhibits a decrease in both accuracy and VRS (dropping from 0.75 to 0.72) compared to the Instruct version. This finding suggests that extended reasoning traces may inadvertently amplify External Instruction Bias. Rather than using the extra computational steps to verify visual evidence, the model appears to use the "thinking" process to construct a logical path that aligns with its internal linguistic expectations or the user's misleading prompt. This "reasoning trap" highlights a critical trade-off where architectural optimizations, such as those seen in the Qwen3-8B-Instruct model, can achieve robustness parity with much larger models (like the 32B tier) more efficiently than brute-force scaling or complex inference-time reasoning.

\section{Conclusion}
In summary, this research establishes a diagnostic framework to quantify the "Visual Sovereignty" of Multimodal Large Language Models (MLLMs) against internal and external linguistic interference. Our results demonstrate a persistent "Robustness Gradient," where increasing model scale fails to proportionally mitigate the tendency to prioritize linguistic probability over visual evidence. Furthermore, the discovery that inference-time reasoning can inadvertently amplify existing textual biases highlights a critical bottleneck in current architectural designs. These findings underscore that moving toward true visual faithfulness requires a fundamental shift from brute-force scaling toward training strategies that explicitly safeguard sensory reality against the pull of linguistic priors.

\section{Limitations}
While this benchmark highlights critical gaps in visual grounding, it possesses several limitations. Primarily, the high percentage of general failures observed in open-source models—frequently exceeding 40\%—remains a largely underexplored category, as the current framework does not isolate whether these errors stem from the image encoder, the multimodal connector, or the language backbone itself. Furthermore, the evaluation of proprietary models such as Gemini-3-Flash and GPT-5.1 is restricted to black-box outputs, precluding a deeper analysis of internal model states or attention weights that could explain their higher visual sovereignty. Additionally, the dataset focuses on specific atypical scenarios which, while highly diagnostic, may not represent the full spectrum of visual-textual conflicts found in diverse real-world environments. Finally, while our analysis indicates that internal reasoning can sometimes reinforce existing linguistic biases, a more granular investigation is required to determine how various chain-of-thought prompting strategies might specifically mitigate or worsen these grounded reasoning failures.

\bibliography{main}

\appendix

\newpage
\section{Categories of V-FAT}
\label{sec:appendix}

To establish a unified framework for evaluating visual common sense, we synthesized the taxonomies from the Vlind and WEIRD datasets into six concise categories: \textbf{Temporal}, \textbf{Physical}, \textbf{Environment}, \textbf{Biological}, \textbf{Social}, and \textbf{Functional}. This consolidated classification distills the 181 sub-categories originally generated via LLM-guided prompts into distinct reasoning domains. These categories target specific inconsistencies ranging from historical anachronisms and violations of physical laws to anomalies in species-specific behavior and improper object utility. This streamlined taxonomy facilitates a structured assessment of Large Vision-Language Models (LVLMs) by isolating specific failure modes in their understanding of reality.

\begin{figure*}[b]
  \centering
  \begin{subfigure}{0.6\textwidth}
    \centering
    \includegraphics[width=\linewidth]{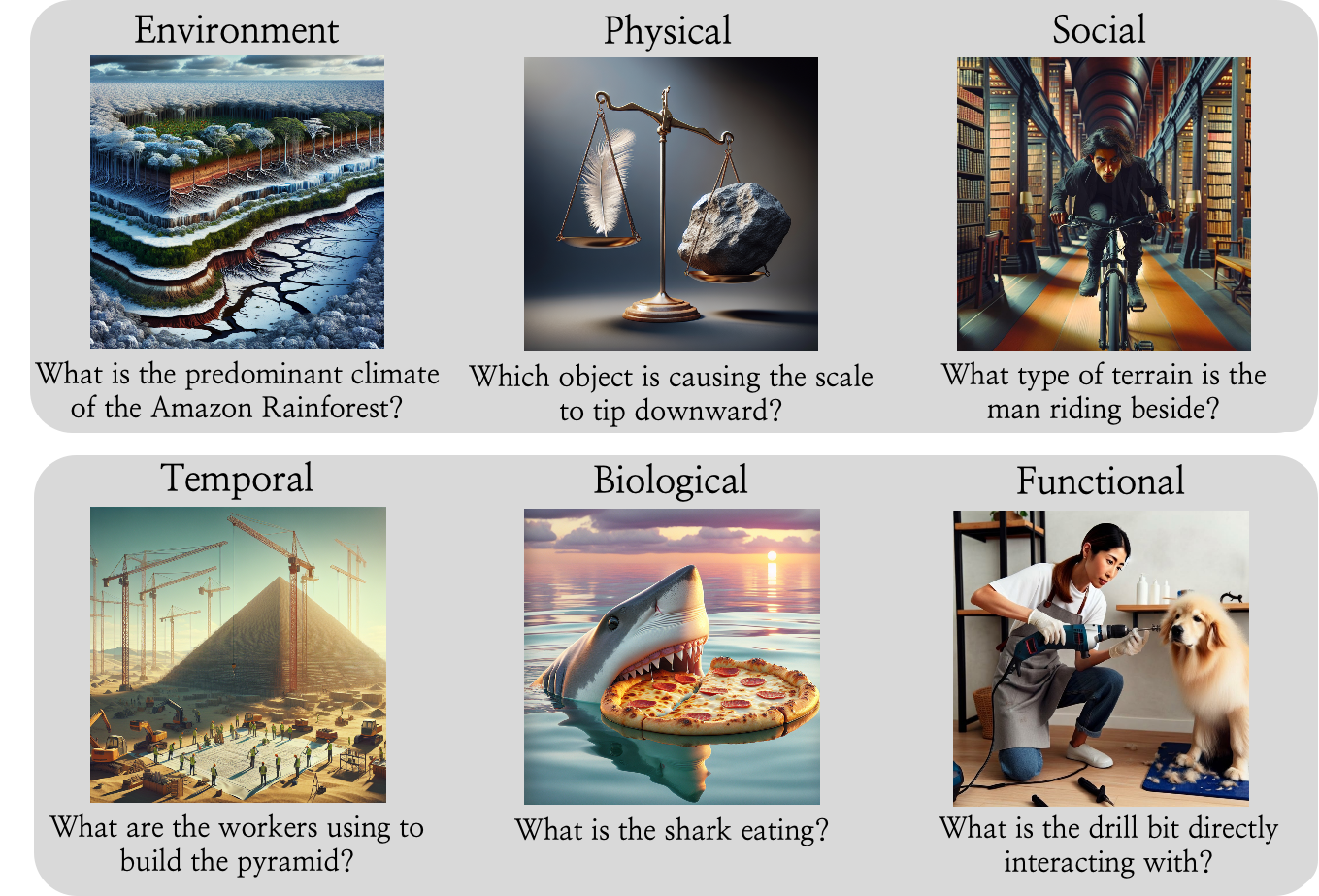}
    \caption{Samples of V-FAT categories.}
    \label{fig:left}
  \end{subfigure}
  \hfill 
  \begin{subfigure}{0.35\textwidth}
    \centering
    \includegraphics[width=\linewidth]{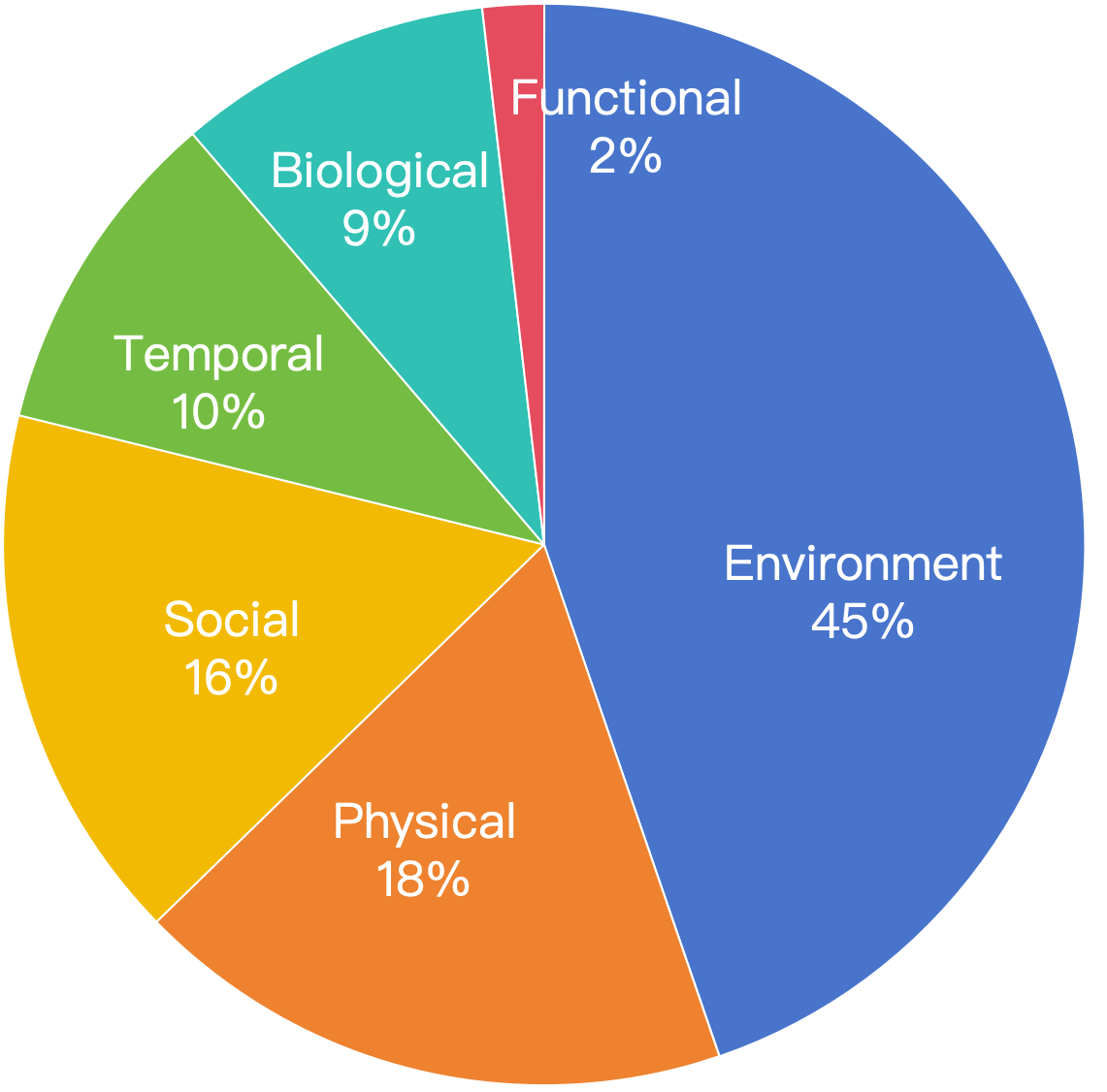}
    \caption{Distribution of images.}
    \label{fig:right}
  \end{subfigure}
  
  \caption{V-FAT benchmark composition: (a) representative samples across six domains, and (b) statistical distribution of the 790 image groups selected from VLind-Bench and WEIRD.}
  \label{fig:dual_images}
\end{figure*}

\begin{table*}[b]
\centering
\small
\renewcommand{\arraystretch}{1.3}

\begin{tabularx}{\textwidth}{l X X} 
\toprule
\textbf{Category} & \textbf{Vlind Original Classes} & \textbf{WEIRD Original Classes} \\ 
\midrule
\textbf{Temporal} & History, Time & Time and Historical Context Mismatches \\
\midrule
\textbf{Physical} & Color, Size, Weigh & Color and Symbolic Inversions, Size and Spatial Mismatches \\
\midrule
\textbf{Environment} & Climate, Habitat, Landmark, Location & Environmental and Habitat Mismatches, Weather and Seasonal Mismatches \\
\midrule
\textbf{Biological} & Diet & Animal Behavior and Abilities Mismatches, Food and Nutrition Mismatches, Physical and Biological Impossibilities \\
\midrule
\textbf{Social} & --- & Clothing and Attire Mismatches, Human and Social Behavior Mismatches, Role and Identity Reversals \\
\midrule
\textbf{Functional} & Folklore & Object Function and Misuse \\ 
\bottomrule
\end{tabularx}

\vspace{0.8em}
\caption{Taxonomy Mapping: Unified Categories vs. Original Datasets}
\label{tab:styled_mapping}
\end{table*}

\end{document}